\documentclass[10pt,twocolumn,letterpaper]{article}

\usepackage[pagenumbers]{cvpr}

\usepackage{subcaption}
\usepackage{graphicx}
\usepackage{booktabs}
\usepackage{amsfonts}
\usepackage{amsmath}
\usepackage{amssymb}
\usepackage[table]{xcolor}
\usepackage{pifont}
\usepackage{diagbox}
\usepackage{multirow}

\definecolor{lightblue}{RGB}{230,242,255}

\definecolor{cvprblue}{rgb}{0.21,0.49,0.74}
\usepackage[pagebackref,breaklinks,colorlinks,allcolors=cvprblue]{hyperref}

\title{DF-CBM: Region-Aware Concept Bottleneck Models for Deepfake Detection}

\author{
Georgios Tsoumplekas$^{1,*}$,
Vazgken Vanian$^{2}$,
Alexandros Doumanoglou$^{2}$,
Panos K. Papadopoulos$^{2}$, \\
Yannis Spyridis$^{3}$,
Dimitrios Zarpalas$^{2}$,
Vasileios Argyriou$^{1}$ \\[4pt]
$^{1}$ Department of Networks and Digital Media, Kingston University London, UK \\
$^{2}$ Centre for Research and Technology Hellas (CERTH), Thessaloniki, Greece \\
$^{3}$ Department of Computer Science, Kingston University London, UK \\[4pt]
$^{*}$ Corresponding author: \texttt{Giorgos.Tsoumplekas@kingston.ac.uk}
}

\begin{document}
\maketitle

\begin{abstract}
Deepfake detection methods have become increasingly effective yet most provide limited insight into the evidence behind their predictions. However, in forensic settings users also need to know which manipulation cues support the decision and where they appear. Existing explainability methods only partially address this need since localization-based approaches lack semantic descriptions while language-based explanation methods are only weakly grounded in visual evidence. In this work, we propose DF-CBM, a region-aware concept bottleneck model for explainable deepfake detection. DF-CBM builds a compact vocabulary of manipulation-related concepts from textual artifact annotations and links each concept to plausible facial and boundary regions. It then predicts these concepts from visual features using a concept-specific masked attention mechanism guided by parsed facial masks and the final real/fake decision is made from the predicted concept bottleneck. Our experiments show that DF-CBM outperforms concept-based baselines in concept prediction and deepfake classification while remaining competitive with state-of-the-art black-box detectors. Finally, qualitative results and intervention analyses demonstrate that DF-CBM provides spatially grounded concept evidence and enables counterfactual explanations of how individual manipulation concepts influence the final prediction. Our code is available at: \href{https://github.com/GeorgeTsoumplekas/DF-CBM}{https://github.com/GeorgeTsoumplekas/DF-CBM}.
\end{abstract}    

\section{Introduction}
\label{sec:introduction}

Recent advances in generative modeling~\cite{esser2024scaling, karras2018progressive, rombach2022high, tian2024visual, yang2025cogvideox} have enabled the synthesis and manipulation of highly realistic visual media commonly referred to as deepfakes~\cite{mirsky2021creation}. While these technologies have beneficial applications in entertainment~\cite{cui2025hallo3, li2026personalive}, media production~\cite{zhou2026autocut} and digital art~\cite{guo2024animatediff}, they also create substantial opportunities for misuse including political disinformation, identity fraud, coordinated misinformation campaigns and non-consensual content generation~\cite{kirchengast2020deepfakes, vaccari2020deepfakes}.

To address these risks, a wide range of deepfake detection methods have been developed to distinguish authentic from manipulated content~\cite{rossler2019faceforensics, li2020face, shiohara2022detecting, cui2025forensics, sun2026dfd}. Yet, despite substantial progress, many detectors remain limited by their black-box nature providing little insight into the visual evidence that drives their predictions. This lack of transparency is particularly problematic in forensic and high-stakes settings where reliable decision-making requires not only accurate predictions but also explanations that are transparent, verifiable and suitable for human oversight.

\begin{figure*}[t]
    \centering
    \includegraphics[width=\textwidth]{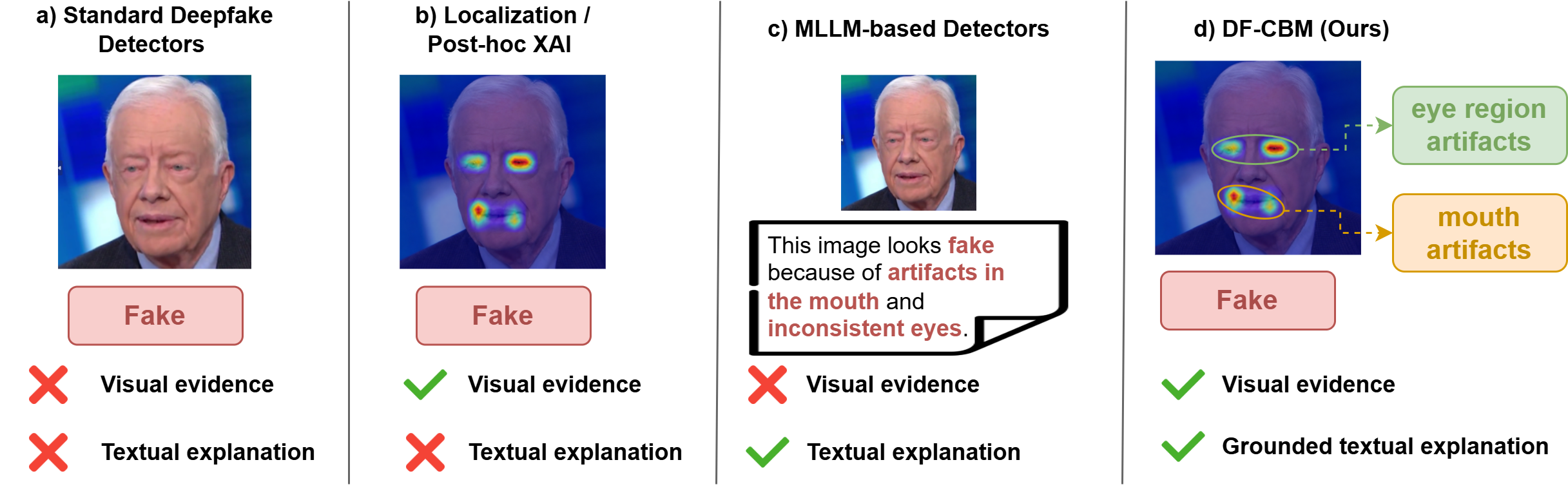}
    \caption{Explanation capabilities of existing deepfake detection approaches and DF-CBM that provides both localized visual evidence and grounded manipulation concepts.}
    \label{fig:introduction_figure}
\end{figure*}

Existing explainable deepfake detection methods mainly follow two directions, as illustrated in Fig.~\ref{fig:introduction_figure}. Localization-based approaches, such as saliency maps, attention visualizations and other post-hoc explanations, highlight image regions that influence a detector's prediction~\cite{bouter2024protoexplorer, soltandoost2025extracting, vanian2026fake}. However, these methods typically indicate where the model focuses without providing a semantic description of the forensic evidence. In contrast, recent MLLM-based approaches generate textual explanations that describe possible manipulation artifacts in natural language~\cite{sun2025towards, zhang2024common, huang2025sida, xu2025fakeshield}. While more interpretable to humans, such explanations are often only weakly grounded in the image and may describe artifacts without establishing a precise correspondence to the supporting visual evidence. As a result, current methods either provide visual evidence without semantic concepts, or textual explanations without sufficiently reliable spatial grounding.

To address this gap, we introduce DF-CBM, a visually grounded concept bottleneck framework for explainable deepfake detection. DF-CBM first constructs a compact vocabulary of manipulation-related concepts and associates each concept with plausible facial or boundary regions. It then learns a region-aware concept bottleneck model that predicts these concepts from visual features while encouraging each concept to rely on the facial regions where that artifact is expected to appear. This way, each predicted manipulation concept is associated not only with a semantic label but also with localized visual evidence, improving the interpretability of the final deepfake prediction.

Notably, unlike conventional black-box detectors and post-hoc explanation methods, DF-CBM makes the intermediate forensic evidence explicit since the final prediction is obtained from predicted manipulation concepts rather than directly from unconstrained visual features. This enables counterfactual analyses and allows users to inspect which concepts contributed to the decision and where their supporting evidence appears in the image. We evaluate DF-CBM under intra-dataset and cross-dataset settings, comparing it against both state-of-the-art black-box detectors and concept-based baselines. Our results show that DF-CBM outperforms concept-based baselines in both concept prediction and real/fake classification while achieving competitive detection performance against state-of-the-art detectors. We further provide qualitative attention maps and contribution scores to visualize the spatial evidence and decision-relevant concepts and conduct a concept intervention analysis to examine how modifying individual concept activations affects the final prediction. Our contributions can be summarized as follows:

\begin{itemize}
    \item We introduce DF-CBM, a region-aware concept bottleneck framework for explainable deepfake detection that predicts semantically meaningful manipulation concepts and grounds them in anatomically relevant facial regions.

    \item We develop an automated pipeline for constructing and grounding forensic concepts, combining text-based concept extraction with facial parsing masks and concept-specific attention to support each concept prediction with localized visual evidence.

    \item We demonstrate that DF-CBM outperforms concept-based baselines in both concept prediction and deepfake detection while achieving competitive detection performance against state-of-the-art black-box detectors and enhanced interpretability through attention maps, contribution scores and concept intervention analysis.

\end{itemize}

\section{Related Work}
\label{sec:related_work}

\subsection{Deepfake Detection Methods}

Early deepfake detection methods focused on identifying visual inconsistencies introduced by face manipulation pipelines including physiological, geometric and boundary-related artifacts~\cite{yang2019exposing, li2020face}. With the release of large-scale benchmarks~\cite{rossler2019faceforensics, li2020celeb, dolhansky2020deepfake, jiang2020deeperforensics, zhou2021face} CNN-based detectors became the dominant paradigm. Several influential methods observed that face manipulation often introduces blending artifacts near facial boundaries leading to approaches such as Face X-Ray~\cite{li2020face} and Self-Blended Images~\cite{shiohara2022detecting} which improve generalization by learning manipulation-independent cues. 

More recent work has shifted toward generalizable detection under unseen manipulation methods and datasets. UCF~\cite{yan2023ucf} disentangles common forgery cues from content- and method-specific artifacts, while LSDA~\cite{yan2024transcending} expands the latent forgery space to reduce overfitting to known manipulation patterns. Other approaches improve robustness through forgery augmentation~\cite{lin2024fake}, discrepancy learning~\cite{yang2025d}, frequency debiasing~\cite{kashiani2025freqdebias}, frequency-guided adaptation~\cite{badr2026frld} or adaptation of pretrained vision-language representations~\cite{cui2025forensics, koutlis2024leveraging, sun2026dfd}. 

\subsection{Explainability in Deepfake Detection}

Explainability in deepfake detection has been explored through visual evidence, prototype-based reasoning and localized explanation methods. Dynamic Prototype Networks~\cite{trinh2021interpretable} use temporal prototypes to explain video-level deepfake dynamics, while ProtoExplorer~\cite{bouter2024protoexplorer} provides a visual analytics interface for prototype-based forensic models. Other works localize decision-relevant facial evidence using local explanations or post-hoc analysis~\cite{soltandoost2025extracting, vanian2026fake}. While such methods improve transparency over standard black-box detectors, they often explain predictions through saliency, prototypes or discovered concepts that are not explicitly constrained to correspond to predefined, human-interpretable manipulation concepts grounded in facial anatomy.

A complementary line of work uses language supervision or multimodal large models to generate richer explanations of manipulated content. Specifically, Visual-linguistic face forgery detection~\cite{sun2025towards}, DD-VQA~\cite{zhang2024common} and ExDDV~\cite{hondru2026exddv} introduce textual artifact descriptions, question-answering formulations and localized language annotations for explainable deepfake detection. Meanwhile, more recent methods and benchmarks~\cite{huang2025sida, xu2025fakeshield, jiang2026tridf, tan2026veritas, jung2026rich} evaluate or exploit multimodal reasoning for detection, localization and explanation. However, free-form textual explanations may be weakly grounded in the image and can be difficult to verify spatially. In contrast, DF-CBM avoids unconstrained explanation generation by predicting a fixed set of manipulation concepts, each explicitly linked to anatomically meaningful facial regions.

\subsection{Concept Bottleneck Models}

Concept Bottleneck Models (CBMs) provide ante-hoc interpretability by constraining predictions to pass through an intermediate layer of human-understandable concepts~\cite{koh2020concept, knab2026whats}. This enables concept-level explanations and test-time interventions but early CBMs required dense concept annotations, limiting their scalability. Subsequent work has relaxed this requirement through richer concept representations~\cite{espinosa2022concept} and weaker supervision~\cite{wang2023learning}.

Recent CBM variants have further improved scalability and spatial grounding. Label-Free CBMs~\cite{oikarinen2023labelfree} and Post-hoc CBMs~\cite{yuksekgonul2023posthoc} leverage vision-language models and concept activation vectors to construct interpretable bottlenecks without dense manual annotations. Spatially-aware approaches, such as SALF-CBM~\cite{benou2025show} and locality-aware CBMs~\cite{jeon2024localityaware} extend CBMs from global concept activations to localized concept maps while CLIP-free extensions broaden their applicability beyond vision-language supervision~\cite{sammani2026clip}. Yet despite this progress, CBMs have not been systematically explored for explainable deepfake detection, where concepts must capture subtle manipulation artifacts and be grounded in anatomically meaningful facial regions.

\section{Methodology}
\label{sec:methodology}

Given an input face image, DF-CBM predicts a set of semantically meaningful manipulation concepts and uses them as an intermediate representation for final real/fake classification. The framework follows a two-stage pipeline. In the first stage, we construct a compact vocabulary of manipulation-related concepts from text-enhanced deepfake datasets and associate each concept with plausible facial or boundary regions using large language models (LLMs). In the second stage, we train a region-aware concept bottleneck model to predict these concepts from frozen visual features while explicitly grounding each concept prediction in the corresponding facial regions.

\begin{figure*}[t]
    \centering
    \includegraphics[width=\textwidth]{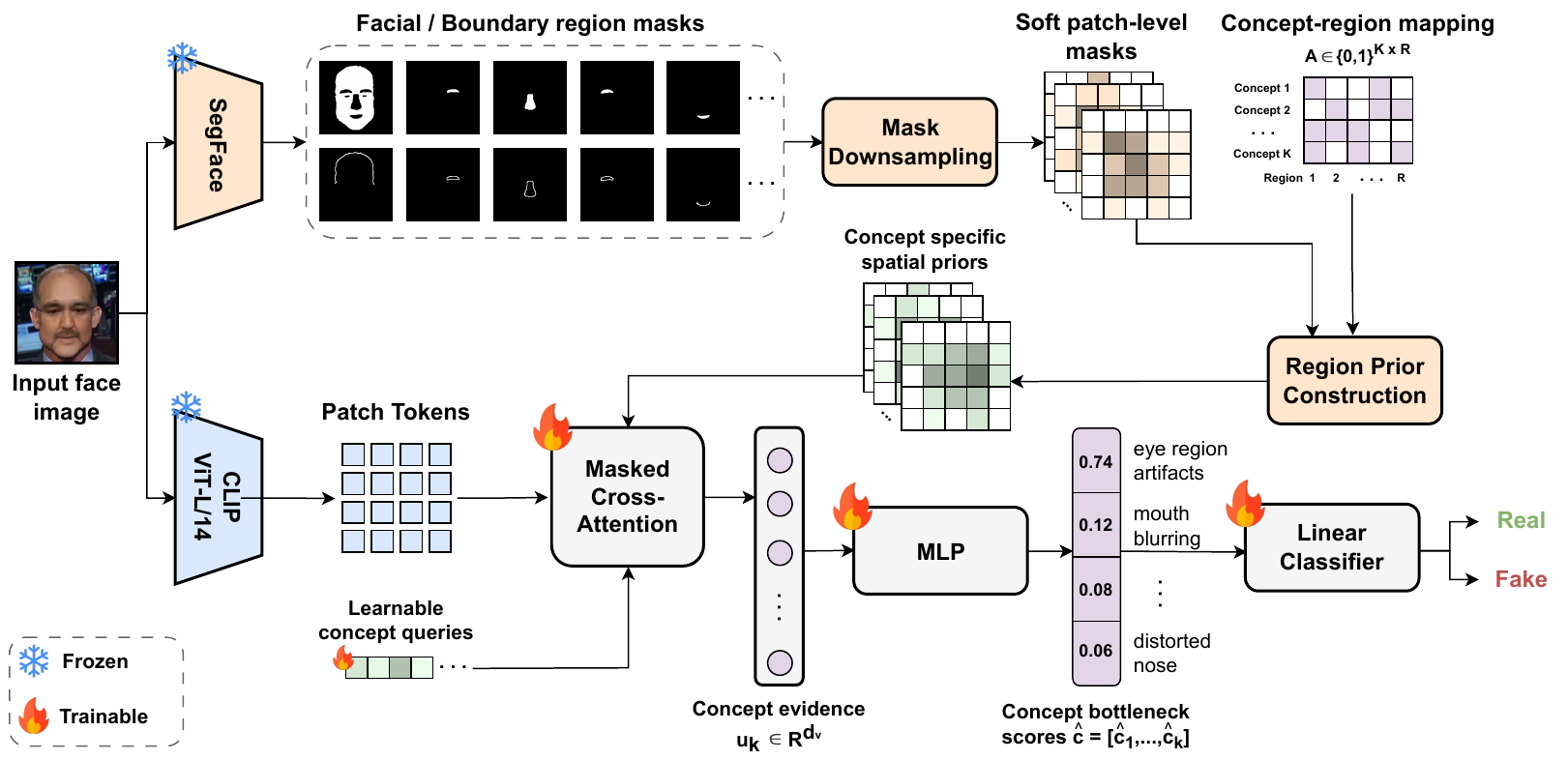}
    \caption{Overview of the region-aware concept bottleneck model in DF-CBM.}
    \label{fig:df_cbm_architecture}
\end{figure*}

\subsection{Concept Vocabulary Extraction and Region Mapping}
~\label{sec:concept_extraction}

The first stage of DF-CBM aims to construct a compact set of manipulation-related concepts and associate each concept with the facial regions in which is expected to appear. This produces a region-grounded concept vocabulary that is later used to guide the region-aware concept bottleneck.

We start from textual annotations in deepfake explanation datasets, including ExDDV~\cite{hondru2026exddv}, DD-VQA~\cite{zhang2024common} and FakeClue~\cite{wen2026spot}. These annotations describe visual evidence of manipulation types in natural language. Given each text description annotation, we prompt Qwen3.5-9B~\cite{yang2025qwen3} to extract atomic manipulation concepts that are concise visual cues that correspond to individual artifacts rather than full sentence-level explanations. The goal is to decompose complex descriptions into simpler concept candidates, such as eye artifacts, mouth distortions, boundary inconsistencies and global facial abnormalities.

The initial concept set is then filtered and consolidated to remove noisy and redundant entries. We first deduplicate the extracted concepts and encode them using EmbeddingGemma-300M~\cite{vera2025embeddinggemma}. The resulting embeddings are clustered using HDBSCAN to group semantically equivalent concepts. For each cluster, we select the top-$k$ entries closest to the cluster centroid and provide them to the same LLM which synthesizes a standardized concept label. The same top-$k$ entries and concept label are then used by the LLM to assign each concept to a predefined set of facial and boundary regions, corresponding to the regions extracted by the facial parsing model in Sec.~\ref{sec:region_aware_concept_bottleneck}. This process yields a binary concept--region mapping $A \in \{0,1\}^{K\times R}$, where $K$ is the number of concepts and $R$ is the number of facial and boundary regions. Each entry $A_{k,r}=1$ indicates that region $r$ is anatomically relevant for concept $k$ and may be used as visual evidence by the bottleneck model. 

Finally, we remove rare concepts that occur in fewer than $N_{\min}$ training samples, with $N_{\min}=750$ in our experiments. This filtering avoids training concept predictors for poorly supported concepts and improves the reliability of the final vocabulary. The vocabulary size is therefore governed by a trade-off between semantic granularity and supervisory density. Here, we favor a compact set of recurring manipulation cues that are sufficiently represented in the data, improving concept prediction robustness while preserving interpretability. After applying this criterion, the resulting vocabulary contains $K=6$ concepts and serves as the semantic basis for the second stage of DF-CBM. Importantly, the architecture does not impose a fixed vocabulary size, allowing larger concept sets when sufficiently dense and reliable supervision is available.

\subsection{Region-Aware Concept Bottleneck Model}
\label{sec:region_aware_concept_bottleneck}

Fig.~\ref{fig:df_cbm_architecture} illustrates the second stage of DF-CBM, which predicts manipulation-related concepts while grounding each prediction in anatomically relevant facial regions. Specifically, given a face image $x \in \mathbb{R}^{3 \times H \times W}$ with height $H$ and width $W$, the model outputs a concept bottleneck score $\hat{\mathbf{c}}=[\hat{c}_1,\ldots,\hat{c}_K]$, where each $\hat{c}_k \in [0,1]$ denotes the predicted presence probability of concept $k$.

We first extract patch-level visual features using a frozen CLIP~\cite{radford2021learning} image encoder. This produces a sequence of patch tokens $F=\{f_i\}_{i=1}^N$, with $f_i \in \mathbb{R}^{D}$, where $N=256$ corresponds to the $16 \times 16$ patch grid and $D$ denotes the CLIP feature dimensionality. In parallel, we use a frozen SegFace~\cite{narayan2025segface} facial parsing model to obtain semantic facial region masks at the same image resolution. For each facial region $r$, we denote the corresponding binary mask as $M^{\mathrm{img}}_r \in \{0,1\}^{H \times W}$.

In addition to the semantic facial regions, we derive boundary masks between predefined adjacent regions, such as skin--nose, skin--hair, mouth--upper lip and lower lip--upper lip. The goal of these masks is to capture transition areas between facial parts motivated by the fact that deepfake artifacts often appear near region transitions, blending seams or facial part boundaries~\cite{li2020face}. Boundary masks are computed directly from the facial parsing output by selecting pixels lying on the boundary of two neighboring regions.

To align the facial masks with the CLIP patch tokens, each image-level mask $M_r^{\mathrm{img}}$ is downsampled to the $16 \times 16$ patch grid using average pooling with kernel size and stride equal to the CLIP patch size. This gives a soft patch-level mask $M_r=[m_{r,1},\ldots,m_{r,N}]$, where $m_{r,i}\in[0,1]$ denotes the fraction of patch $i$ covered by region $r$. The use of soft masks avoids assigning each patch to a single region and preserves partial region coverage within each patch.

Next, for each concept $k$, we construct a concept-specific spatial prior $P_k=[p_{k,1},\ldots,p_{k,N}]$ over the CLIP patch grid by combining the patch-level masks of the regions associated with that concept:

\begin{equation}
    p_{k,i} = \max_{r:A_{k,r}=1} m_{r,i}.
\end{equation}

Here, $A_{k,r}=1$ is the concept-region mapping entry indicating that region $r$ is relevant for concept $k$. We use the maximum so that a patch receives high prior weight if it overlaps with any region associated with the concept. This is done because many manipulation concepts may correspond to multiple plausible facial regions and averaging would unnecessarily weaken the prior when only one of these regions is present in a patch. Overall, the resulting concept-specific spatial prior specifies which patches are anatomically plausible evidence for predicting concept $k$.

However, the actual evidence for each concept must still be inferred from the visual features. We therefore use a concept-specific masked cross-attention mechanism to extract visual evidence from the CLIP patch tokens. Each concept $k$ is represented by a learnable query vector $q_k \in \mathbb{R}^{d}$, initialized from the CLIP text embedding of the corresponding concept name. The query attends over the patch tokens while the spatial prior is added to the attention logits as a soft mask:

\begin{equation}
    \alpha_{k,i} =
    \operatorname{softmax}_{i}
    \left(
    \frac{q_k^{\top} W_K f_i}{\sqrt{d}}
    + \log(p_{k,i}+\epsilon)
    \right),
\end{equation}

where $W_K \in \mathbb{R}^{d \times D}$ projects CLIP features into the query space, $p_{k,i}\in[0,1]$ is the spatial prior value for concept $k$ at patch $i$, $d$ is the attention dimension and $\epsilon>0$ is a small constant for numerical stability. The logarithmic prior acts as a soft masking mechanism that suppresses patches outside the relevant facial regions while still allowing soft attention within the valid support.

Using the resulting attention weights, we aggregate the value-projected patch features to obtain a concept-specific visual evidence vector:

\begin{equation}
    u_k = \sum_{i=1}^{N} \alpha_{k,i} W_V f_i,
\end{equation}

where $W_V \in \mathbb{R}^{d_v \times D}$ is a learned value projection and $u_k \in \mathbb{R}^{d_v}$. The vector $u_k$ summarizes the visual evidence relevant to concept $k$ by pooling information from the CLIP patches according to the concept-specific attention mechanism. Thus, different concepts can extract different evidence from the same image even though they share the same underlying patch tokens.

The evidence vector is then mapped to a shared concept feature space using:

\begin{equation}
    h_k = \operatorname{ReLU}(W_s u_k + b_s),
\end{equation}

where $W_s \in \mathbb{R}^{d_h \times d_v}$ and $b_s \in \mathbb{R}^{d_h}$ are shared across all concepts and $h_k \in \mathbb{R}^{d_h}$. A concept-specific prediction head then maps $h_k$ to a scalar concept probability:

\begin{equation}
    \hat{c}_k = \sigma(w_k^\top h_k + b_k),
\end{equation}

where $w_k \in \mathbb{R}^{d_h}$ and $b_k \in \mathbb{R}$ are concept-specific parameters and $\hat{c}_k \in [0,1]$ denotes the predicted probability that concept $k$ is present. Overall, this formulation also yields three interpretable outputs: the concept--region mapping $A$ specifies where each concept can look, the attention weights $\alpha_{k,i}$ localize the supporting patches and the bottleneck scores $\hat{c}_k$ indicate which manipulation concepts are present.

Finally, the concept bottleneck scores are used for real/fake prediction through a linear classifier:

\begin{equation}
    z = w_{\mathrm{cls}}^\top \hat{\mathbf{c}} + b_{\mathrm{cls}},
\end{equation}

where $w_{\mathrm{cls}}\in\mathbb{R}^{K}$ and $b_{\mathrm{cls}}\in\mathbb{R}$ are the classifier parameters and $z\in\mathbb{R}$ is the binary classification logit. Since the classifier operates directly on the concept bottleneck, each classifier weight links a concept activation to the final detection decision.

\subsection{Model Training} ~\label{sec:model_training}

Given an input image $x$, DF-CBM outputs a concept bottleneck score $\hat{\mathbf{c}}(x)= [\hat{c}_{1}(x),\ldots,\hat{c}_{K}(x)]$, where $\hat{c}_k(x)\in[0,1]$ denotes the predicted probability of concept $k$ and classification logits $z(x)$ for real/fake prediction. During training, each image is associated with a binary label $y\in\{0,1\}$ where $y=1$ denotes a fake image and a concept target encoded in a binary vector $\mathbf{c}\in\{0,1\}^{K}$. We train the concept bottleneck using a weighted binary cross-entropy loss:

\begin{equation}
    \mathcal{L}_{\mathrm{con}}(x)
    =
    \frac{1}{K}
    \sum_{k=1}^{K}
    \omega_k(x)\,
    \mathrm{BCE}(\hat{c}_k(x), c_k),
\end{equation}

where $c_k$ is the target label for concept $k$ and $w_k(x)$ is defined as:

\begin{equation}
    \omega_k(x) =
    \begin{cases}
        \rho_k, & c_k=1,\\[2pt]
        1, & c_k=0,\; y=0,\\[2pt]
        \beta, & c_k=0,\; y=1.
    \end{cases}
    \label{eq:noise_weight}
\end{equation}

Here, $\rho_k=N_k^{-}/N_k^{+}$ is a concept-specific positive weight computed from the training set, where $N_k^{+}$ and $N_k^{-}$ denote the number of positive and negative labels for concept $k$, respectively, used to account for the strong imbalance in concept annotations. The parameter $\beta\in[0,1]$ down-weights negative concept labels on fake images to reflect asymmetric label reliability. Positive labels and negatives from real images are treated as reliable, whereas negative concept labels in fake images may correspond to unannotated artifacts. We therefore treat these entries as weak negatives, reducing their influence while preserving useful supervision.

The real/fake classifier is trained with a standard cross-entropy loss $\mathcal{L}_{\mathrm{cls}}(x) = \mathrm{CE}(z(x), y)$ over the classification logits $z(x)$ and the full training objective is:

\begin{equation}
    \mathcal{L}
    =
    \mathbb{E}_{(x,y,\mathbf{c})}
    \left[
    \mathcal{L}_{\mathrm{cls}}(x)
    +
    \lambda \mathcal{L}_{\mathrm{con}}(x)
    \right],
    \label{eq:joint}
\end{equation}

where $\lambda>0$ controls the strength of concept supervision. The region-aware concept bottleneck and the final classifier are optimized jointly under this objective.

\begin{table}[h]
    \centering
    \caption{\textit{Macro-averaged} concept prediction performance on the FaceForensics++ test set.}
    \label{tab:concept_results_average}
    \setlength{\tabcolsep}{3pt}
    \footnotesize
    \resizebox{0.8\linewidth}{!}{
    \begin{tabular}{lcccc}
        \toprule
        \textbf{Model} 
        & \textbf{B-Acc.} 
        & \textbf{$F_1$} 
        & \textbf{F-AUC}
        & \textbf{V-AUC} \\
        \midrule
        Joint-CBM~\cite{koh2020concept} & \textbf{0.687} & 0.310 & 0.739 & 0.753 \\
        BotCL~\cite{wang2023learning} & 0.651 & 0.300 & 0.689 & 0.693 \\
        \midrule
        \rowcolor{lightblue}
        \textbf{DF-CBM} & 0.675 & \textbf{0.563} & \textbf{0.743} & \textbf{0.772} \\
        \bottomrule
    \end{tabular}%
    }
\end{table}

\section{Experimental Results}
\label{sec:experimental_results}

\subsection{Experimental Setup}

We train DF-CBM on the concept-annotated subset of FaceForensics++~\cite{rossler2019faceforensics} constructed using the concept extraction procedure in Section~\ref{sec:concept_extraction}. Following DeepfakeBench~\cite{yan2023deepfakebench}, we use the \texttt{c23} version for all experiments. All images are cropped and resized to $224\times224$. DF-CBM uses CLIP ViT-L/14~\cite{radford2021learning} as the image encoder and SegFace~\cite{narayan2025segface} with a Swin-Base backbone for facial region parsing. The attention and value dimensions, as well as the shared concept-evidence projection, are all set to $768$. The model is trained jointly for $15$ epochs using AdamW~\cite{loshchilov2018decoupled} with a learning rate of $0.0002$, batch size $128$, concept loss weight $\lambda=1.0$ and noise discount factor $\beta=0.15$ on 2 RTX A6000 GPUs.

Following prior work~\cite{yan2025orthogonal, sun2026dfd, cui2025forensics}, we report frame- and video-level AUC for real/fake detection. For concept prediction, we report balanced accuracy (B-Acc.), $F_1$, frame-level AUC (F-AUC) and video-level AUC (V-AUC). We evaluate DF-CBM in a cross-method intra-dataset setting, training on FaceForensics++ and testing on eight unseen manipulation methods from DF40~\cite{yan2024df40} and in a cross-dataset setting using CDF-v2~\cite{li2020celeb}, DFD~\cite{dfd2020}, DFDC~\cite{dfdc2020}, DFDCP~\cite{dolhansky2020deepfake} and UADFV~\cite{li2018ictu}. We compare against the concept-based baselines Joint-CBM~\cite{koh2020concept} and BotCL~\cite{wang2023learning}, trained on the same concept-annotated subset as DF-CBM, as well as state-of-the-art black-box detectors using the reported results from~\cite{sun2026dfd, cui2025forensics}.

\subsection{Concept Prediction Results}

We first evaluate whether the learned bottleneck provides meaningful concept predictions. We compare DF-CBM with Joint-CBM~\cite{koh2020concept} and BotCL~\cite{wang2023learning}, on the FaceForensics++ test set. Evaluation is performed on all real test samples and on fake samples annotated with at least one concept from our vocabulary resulting in approximately $4.5$K real samples and $6.5$K fake samples.

Table~\ref{tab:concept_results_average} reports the macro-averaged results across all concepts in terms of balanced accuracy, $F_1$, F-AUC and V-AUC. DF-CBM outperforms both Joint-CBM and BotCL on $F_1$, F-AUC and V-AUC, achieving the largest gain in macro-$F_1$, with a 25.3\% improvement over Joint-CBM, while maintaining comparable balanced accuracy. Overall, these results show that grounding concept prediction in anatomically relevant facial and boundary regions leads to a more reliable and interpretable concept bottleneck.

\begin{table*}[h]
\centering
\caption{Intra-dataset deepfake detection performance on FaceForensics++ using video-level AUC. DF-CBM is compared with black-box detectors (top-part) and concept-based baselines (bottom part) across eight manipulation methods. Best results among the concept-based methods are shown in \textbf{bold}.}
\label{tab:intra_dataset_video_auc}
\setlength{\tabcolsep}{3pt}
\begin{tabular}{lccccccccc}
\toprule
\textbf{Methods} 
& \textbf{UniFace} 
& \textbf{BlendFace} 
& \textbf{MobSwap} 
& \textbf{e4s} 
& \textbf{FaceDan} 
& \textbf{FSGAN} 
& \textbf{InSwap} 
& \textbf{SimSwap} 
& \textbf{Avg.} \\
\midrule
SBI~\cite{shiohara2022detecting}         & 0.724 & 0.891 & 0.952 & 0.750 & 0.594 & 0.803 & 0.712 & 0.701 & 0.766 \\
UCF~\cite{yan2023ucf}         & 0.831 & 0.827 & 0.950 & 0.731 & 0.862 & 0.937 & 0.809 & 0.647 & 0.824 \\
IID~\cite{huang2023implicit}         & 0.839 & 0.789 & 0.888 & 0.766 & 0.844 & 0.927 & 0.789 & 0.644 & 0.811 \\
LSDA~\cite{yan2024transcending}       & 0.872 & 0.875 & 0.930 & 0.694 & 0.721 & 0.939 & 0.855 & 0.793 & 0.835 \\
ProDet~\cite{cheng2024can}   & 0.908 & 0.929 & 0.975 & 0.771 & 0.747 & 0.928 & 0.837 & 0.844 & 0.867 \\
CDFA~\cite{lin2024fake}       & 0.762 & 0.756 & 0.823 & 0.631 & 0.803 & 0.942 & 0.772 & 0.757 & 0.781 \\
Effort~\cite{yan2025orthogonal}   & 0.962 & 0.873 & 0.953 & 0.983 & 0.926 & 0.957 & 0.936 & 0.926 & 0.940 \\
DFD-HR~\cite{sun2026dfd} & 0.988 & 0.956 & 0.986 & 0.993 & 0.970 & 0.984 & 0.981 & 0.970 & 0.978 \\
\midrule
Joint-CBM~\cite{koh2020concept} & 0.932 & \textbf{0.897} & 0.912 & 0.865 & 0.687 & 0.920 & 0.894 & 0.827 & 0.867 \\
BotCL~\cite{wang2023learning} & \textbf{0.954} & 0.881 & \textbf{0.942} & 0.759 & 0.755 & \textbf{0.936} & \textbf{0.937} & 0.878 & 0.880 \\
\rowcolor{lightblue}
\textbf{DF-CBM} & 0.912 & 0.878 & 0.919 & \textbf{0.968} & \textbf{0.845} & 0.928 & 0.835 & \textbf{0.891} & \textbf{0.897} \\
\bottomrule
\end{tabular}%
\end{table*}

\subsection{Deepfake Detection Results}

\paragraph{Intra-dataset evaluation.}

We first evaluate video-level deepfake detection performance across different manipulation methods based on images from FaceForensics++. Table~\ref{tab:intra_dataset_video_auc} compares DF-CBM with both state-of-the-art black-box detectors and concept-based baselines. For the black-box detectors, we report the results from~\cite{sun2026dfd} under the same evaluation protocol. For the concept-based baselines, Joint-CBM~\cite{koh2020concept} and BotCL~\cite{wang2023learning} are trained on the same concept-annotated subset of FaceForensics++ used to train DF-CBM.

Among concept-based methods, DF-CBM achieves the best average video-level AUC, improving over both Joint-CBM and BotCL. Compared with black-box detectors, DF-CBM remains competitive, achieving the third-best average performance overall. Although the strongest fully black-box detectors still obtain higher AUC, this is expected because DF-CBM constrains the prediction to pass through a compact concept bottleneck. In return, DF-CBM provides explicit concept-level evidence and spatial grounding which are not available in standard black-box detectors.

\paragraph{Cross-dataset evaluation.}

Tables~\ref{tab:cross_dataset_video_auc} and~\ref{tab:cross_dataset_frame_auc} report video-level and frame-level AUC, respectively, when testing the examined methods on unseen datasets. We report black-box detector results from prior work~\cite{sun2026dfd,cui2025forensics}. Notably, DF-CBM substantially outperforms the concept-based baselines across all target datasets. This improved performance is particularly important in the cross-dataset setting, where models must generalize beyond the manipulation methods and data distribution seen during training indicating that region-aware concept prediction provides a more transferable bottleneck than standard concept-based alternatives. Compared with SOTA black-box detectors, DF-CBM remains competitive, although the strongest unconstrained detectors still achieve higher AUC on several datasets due to the inclusion of the interpretable concept bottleneck layer in DF-CBM.

\begin{table}[t]
\centering
\caption{Cross-dataset deepfake detection performance using video-level AUC. Models are trained on FaceForensics++ and evaluated on unseen datasets. Best results among the concept-based methods (bottom part) are shown in \textbf{bold}.}
\label{tab:cross_dataset_video_auc}
\setlength{\tabcolsep}{3pt}
\resizebox{\linewidth}{!}{%
\begin{tabular}{lcccccccc}
\toprule
\textbf{Methods} 
& \textbf{CDF-v2} 
& \textbf{DFD} 
& \textbf{DFDC} 
& \textbf{DFDCP} 
& \textbf{UADFV}
& \textbf{Avg.} \\
\midrule
SBI~\cite{shiohara2022detecting} & 0.886 & 0.827 & 0.717 & 0.848 & - & - \\
UCF~\cite{yan2023ucf} & 0.837 & 0.867 & 0.742 & 0.770 & - & - \\
IID~\cite{huang2023implicit} & 0.838 & 0.939 & 0.700 & 0.689 & - & - \\
LSDA~\cite{yan2024transcending} & 0.875 & 0.881 & 0.701 & 0.812 & - & - \\
ProDet~\cite{cheng2024can} & 0.926 & 0.901 & 0.707 & 0.828 & - & - \\
CDFA~\cite{lin2024fake} & 0.938 & 0.954 & 0.830 & 0.881 & - & - \\
Effort~\cite{yan2025orthogonal} & 0.956 & 0.965 & 0.843 & 0.909 & - & - \\
ForAda~\cite{cui2025forensics} & 0.957 & 0.972 & 0.872 & 0.929 & - & -\\
DFD-HR~\cite{sun2026dfd} & 0.960 & 0.980 & 0.865 & 0.901 & - & - \\
\midrule
Joint-CBM~\cite{koh2020concept} & 0.646 & 0.674 & 0.625 & 0.633 & 0.974 & 0.710 \\
BotCL~\cite{wang2023learning} & 0.746 & 0.789 & 0.699 & 0.676 & 0.953 & 0.773 \\
\rowcolor{lightblue}
\textbf{DF-CBM} & \textbf{0.891} & \textbf{0.927} & \textbf{0.758} & \textbf{0.816} & \textbf{0.995} & \textbf{0.877} \\
\bottomrule
\end{tabular}%
}
\end{table}

\begin{table}[t]
\centering
\caption{Cross-dataset deepfake detection performance using frame-level AUC. Models are trained on FaceForensics++ and evaluated on unseen datasets. Best results among the concept-based methods (bottom part) are shown in \textbf{bold}.}
\label{tab:cross_dataset_frame_auc}
\setlength{\tabcolsep}{3pt}
\resizebox{\linewidth}{!}{%
\begin{tabular}{lcccccc}
\toprule
\textbf{Methods}
& \textbf{CDF-v2}
& \textbf{DFD}
& \textbf{DFDC}
& \textbf{DFDCP}
& \textbf{UADFV}
& \textbf{Avg.} \\
\midrule
SBI~\cite{shiohara2022detecting} & 0.813 & 0.774 & - & 0.799 & - & - \\
UCF~\cite{yan2023ucf} & 0.753 & 0.807 & 0.719 & 0.759 & - & - \\
ED~\cite{ba2024exposing} & 0.864 & - & 0.721 & 0.851 & - & - \\
CFM~\cite{luo2023beyond} & 0.828 & 0.915 & - & 0.758 & - & - \\
FoCus~\cite{tian2024learning} & 0.720 & - & 0.669 & 0.778 & - & - \\
LSDA~\cite{yan2024transcending} & 0.830 & 0.880 & 0.736 & 0.815 & - & - \\
DiffusionFake~\cite{sun2024diffusionfake} & 0.805 & 0.904 & - & 0.810 & - & - \\
MoE-FFD~\cite{kong2025moe} & 0.867 & 0.904 & - & - & - & - \\
Dual-Adapter~\cite{shao2025deepfake} & 0.717 & - & 0.727 & - & - & - \\
UDD~\cite{fu2025exploring} & 0.869 & 0.910 & 0.758 & 0.856 & - & - \\
Effort~\cite{yan2025orthogonal} & 0.901 & 0.923 & 0.798 & - & - & - \\
FFTG~\cite{sun2025towards} & 0.832 & 0.948 & - & - & - & - \\
ForAda~\cite{cui2025forensics} & 0.900 & 0.933 & 0.843 & 0.890 & - & - \\
DFD-HR~\cite{sun2026dfd} & 0.910 & 0.953 & 0.843 & - & - & - \\
\midrule
Joint-CBM~\cite{koh2020concept} & 0.604 & 0.644 & 0.609 & 0.614 & 0.943 & 0.683 \\
BotCL~\cite{wang2023learning} & 0.683 & 0.738 & 0.676 & 0.649 & 0.919 & 0.733 \\
\rowcolor{lightblue}
\textbf{DF-CBM} & \textbf{0.830} & \textbf{0.874} & \textbf{0.732} & \textbf{0.774} & \textbf{0.983} & \textbf{0.839} \\
\bottomrule
\end{tabular}%
}
\end{table}

\subsection{Qualitative Analysis}

To analyze the visual evidence produced by DF-CBM, we visualize concept attention maps. The attention weights $\alpha_{k,i}$ indicate where the model attends when predicting concept $k$ for image $x$. Since attention alone does not quantify the influence of each concept on the final prediction, we also compute a concept contribution score using the linear classifier $s_k(x)=w_{\mathrm{cls},k}\hat{c}_k(x)$ where $w_{\mathrm{cls},k}$ is the classifier weight associated with concept $k$. This score measures how strongly each activated concept supports the fake prediction.

\begin{figure*}[t]
    \centering
    \begin{subfigure}{0.40\linewidth}
        \centering
        \includegraphics[width=\linewidth]{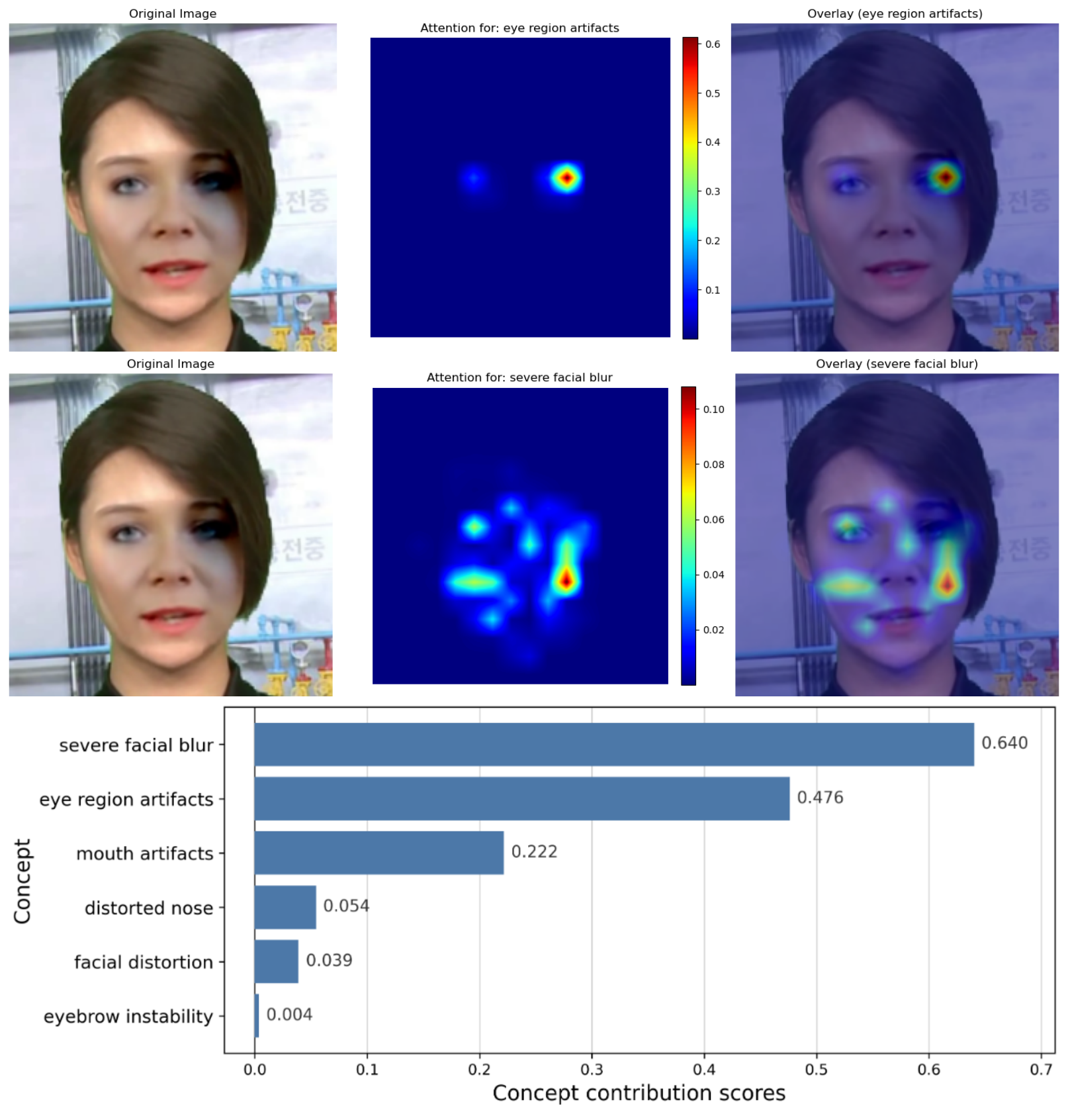}
        \caption{Eye artifacts and facial blur.}
        \label{fig:qualitative_a}
    \end{subfigure}
    \hspace{0.02\linewidth}
    \begin{subfigure}{0.40\linewidth}
        \centering
        \includegraphics[width=\linewidth]{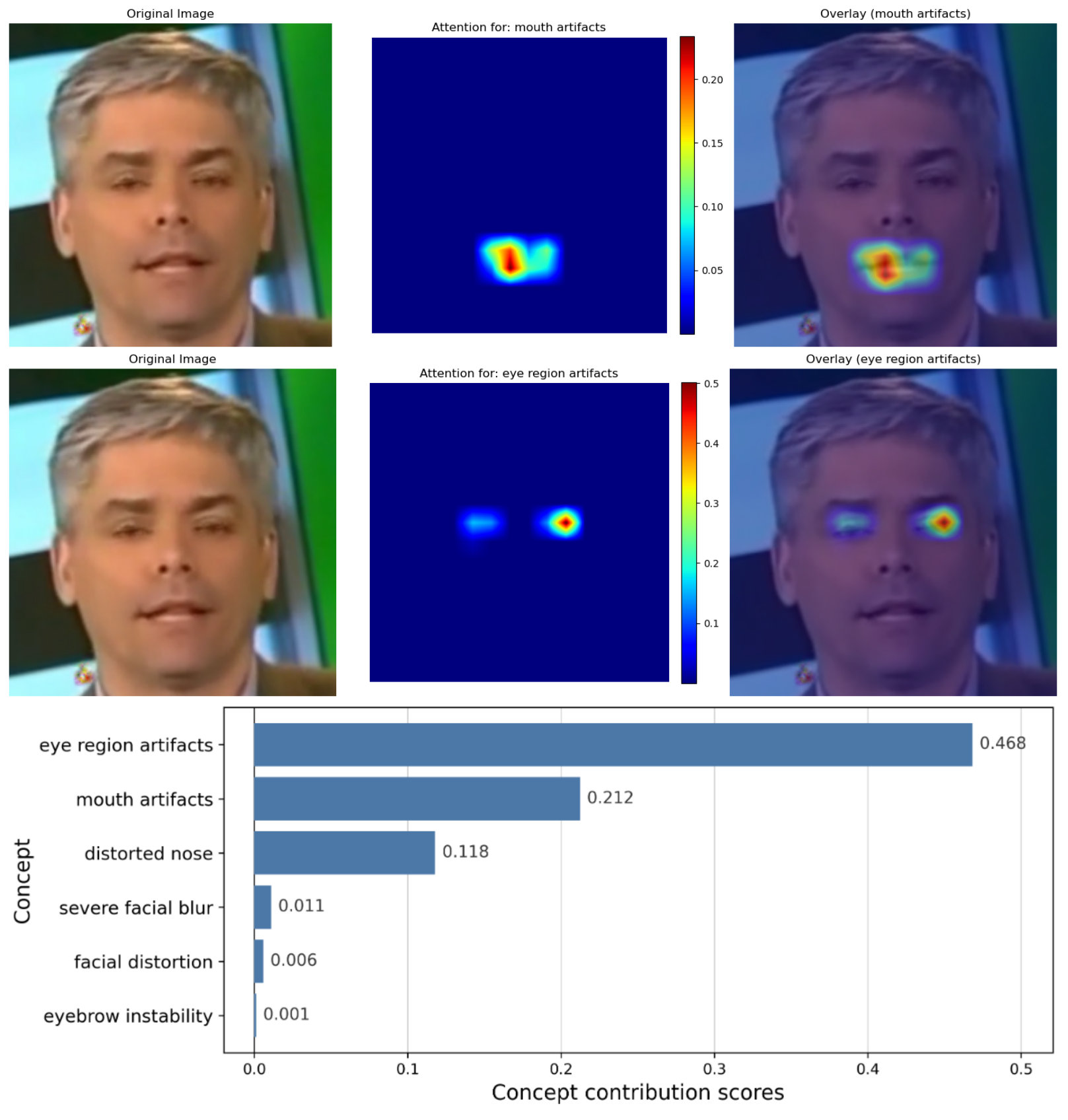}
        \caption{Mouth and eye artifacts.}
        \label{fig:qualitative_b}
    \end{subfigure}
    \caption{Qualitative examples of DF-CBM explanations. For each manipulated image, we show concept-specific attention maps, their overlays on the input image and the corresponding concept contribution scores.}
    \label{fig:qualitative_examples}
\end{figure*}

Fig.~\ref{fig:qualitative_examples} shows two manipulated images containing two annotated manipulation concepts each. In both examples, DF-CBM correctly predicts the annotated concepts which also receive the highest contribution scores. This indicates that the final decision is driven by the relevant manipulation concepts rather than unrelated activations. In Fig.~\ref{fig:qualitative_a}, the model localizes the predicted concepts around the manipulated eye and cheek regions. Similarly, Fig.~\ref{fig:qualitative_b} shows fine-grained localization with higher attention on specific subregions such as the left side of the mouth and the right eye.

\begin{table*}[t]
\centering
\caption{Architectural component ablation on FaceForensics++ using video-level AUC.}
\label{tab:ablation_components}
\setlength{\tabcolsep}{3pt}
\begin{tabular}{lccccccccc}
\toprule
\textbf{Methods} 
& \textbf{UniFace} 
& \textbf{BlendFace} 
& \textbf{MobSwap} 
& \textbf{e4s} 
& \textbf{FaceDan} 
& \textbf{FSGAN} 
& \textbf{InSwap} 
& \textbf{SimSwap} 
& \textbf{Avg.} \\
\midrule
\rowcolor{lightblue}
\textbf{DF-CBM} & \textbf{0.912} & \textbf{0.878} & 0.919 & \textbf{0.968} & 0.845 & \textbf{0.928} & 0.835 & \textbf{0.891} & \textbf{0.897} \\
w/o region prior & 0.884 & 0.789 & 0.884 & 0.926 & 0.812 & 0.879 & 0.807 & 0.809 & 0.849 \\
w/o concept-specific attention & 0.830 & 0.810 & 0.866 & 0.936 & 0.811 & 0.896 & 0.768 & 0.843 & 0.845 \\
w/o concept bottleneck & 0.868 & 0.837 & \textbf{0.927} & 0.933 & \textbf{0.877} & 0.912 & \textbf{0.844} & 0.877 & 0.884 \\
\bottomrule
\end{tabular}%
\end{table*}

\begin{table*}[t]
\centering
\caption{Query design ablation for the masked cross-attention mechanism on FaceForensics++ using video-level AUC.}
\label{tab:ablation_query}
\setlength{\tabcolsep}{3pt}
\begin{tabular}{lccccccccc}
\toprule
\textbf{Methods} 
& \textbf{UniFace} 
& \textbf{BlendFace} 
& \textbf{MobSwap} 
& \textbf{e4s} 
& \textbf{FaceDan} 
& \textbf{FSGAN} 
& \textbf{InSwap} 
& \textbf{SimSwap} 
& \textbf{Avg.} \\
\midrule
\rowcolor{lightblue}
\textbf{DF-CBM} & \textbf{0.912} & \textbf{0.878} & \textbf{0.919} & \textbf{0.968} & 0.845 & 0.928 & \textbf{0.835} & \textbf{0.891} & \textbf{0.897} \\
frozen text queries & 0.911 & 0.867 & \textbf{0.919} & 0.967 & \textbf{0.851} & \textbf{0.932} & 0.829 & 0.887 & 0.895 \\
random query initialization & 0.827 & 0.811 & 0.875 & 0.955 & 0.805 & 0.907 & 0.784 & 0.825 & 0.849 \\
\bottomrule
\end{tabular}%
\end{table*}

\subsection{Ablation Study}

\paragraph{Architectural component ablation.}

Table~\ref{tab:ablation_components} ablates the main architectural components of DF-CBM on the intra-dataset evaluation set. Removing the region prior reduces the average AUC showing that the concept--region mapping provides useful spatial guidance for detecting manipulation concepts. Replacing concept-specific attention further lowers the average AUC to $0.845$ indicating that allowing each concept to extract its own visual evidence is important for both concept prediction and final detection. Removing the concept bottleneck also degrades performance reducing the average AUC to $0.884$ which suggests that explicit manipulation concepts provide useful semantic structure for detection.

\paragraph{Query design ablation.}

We further study the query design used in the masked cross-attention mechanism. Table~\ref{tab:ablation_query} compares randomly initialized learnable queries, frozen queries initialized from CLIP text embeddings and the full DF-CBM with learnable CLIP-initialized queries. The largest gain comes from text-based initialization suggesting that concept semantics help the model associate manipulation concepts with relevant visual evidence while allowing the initialized queries to be further refined during training provides a small additional gain.

\subsection{Interpretability via Intervention Analysis}

\paragraph{Intervention mechanism.}
To evaluate the influence and interpretability of the concept bottleneck layer, we implement a sequential intervention mechanism. For each facial concept, we estimate the empirical distribution of its predicted logits over the training set. To simulate concept activation and suppression while remaining within the range observed during training, intervention values are anchored to extreme quantiles of the corresponding logit distribution. Specifically, concept absence and presence are represented by the $1^{st}$ and $99^{th}$ percentiles, respectively. We target misclassified samples and intervene on one concept at a time, retaining each intervention before applying the next. For false real predictions concepts are sequentially set to their $99^{th}$ percentile intervention values, increasing evidence for deepfake artifacts and steering the prediction towards the \textit{Fake} class. Conversely, for false fake predictions concepts are sequentially set to their $1^{st}$ percentile intervention values, suppressing artifact evidence and shifting the prediction towards the \textit{Real} class.

\begin{figure*}[t]
    \centering
    \begin{subfigure}{0.48\linewidth}
        \centering
        \includegraphics[width=\linewidth]{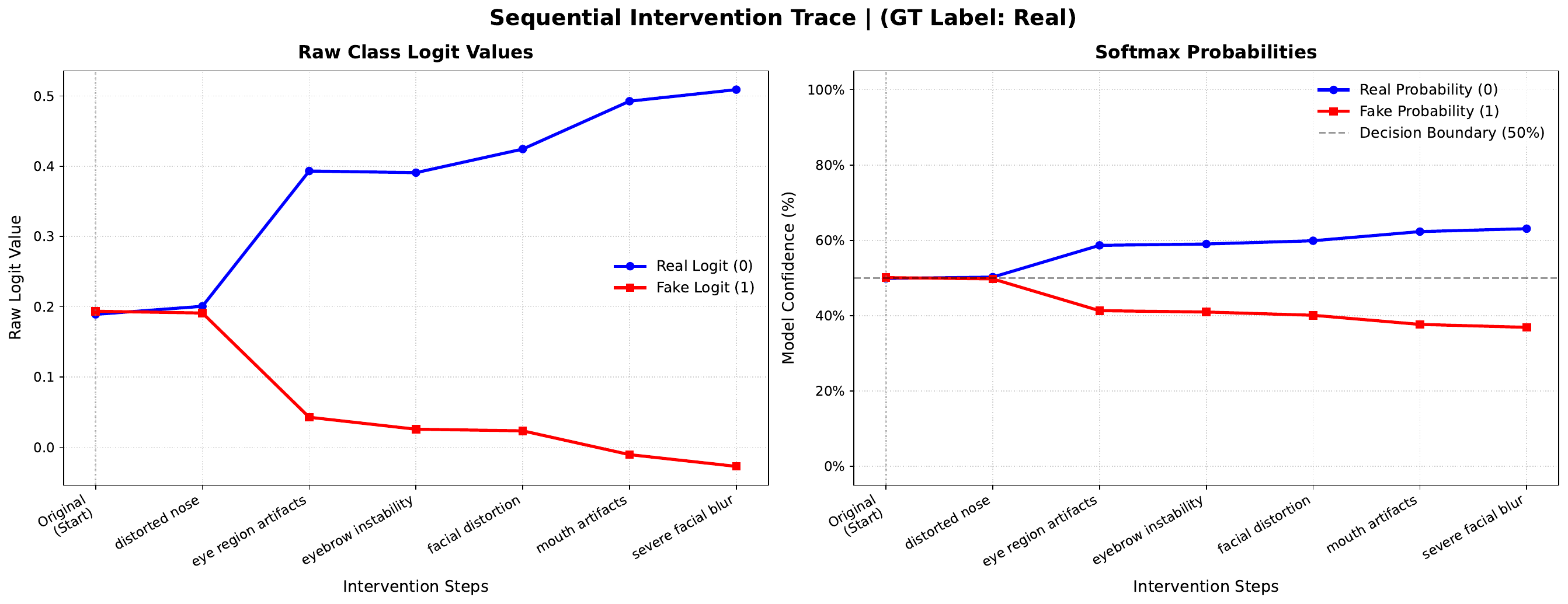}
        \caption{Ground Truth: Real (Predicted: Fake)}
        \label{fig:real_trace}
    \end{subfigure}
    \hspace{0.02\linewidth}
    \begin{subfigure}{0.48\linewidth}
        \centering
        \includegraphics[width=\linewidth]{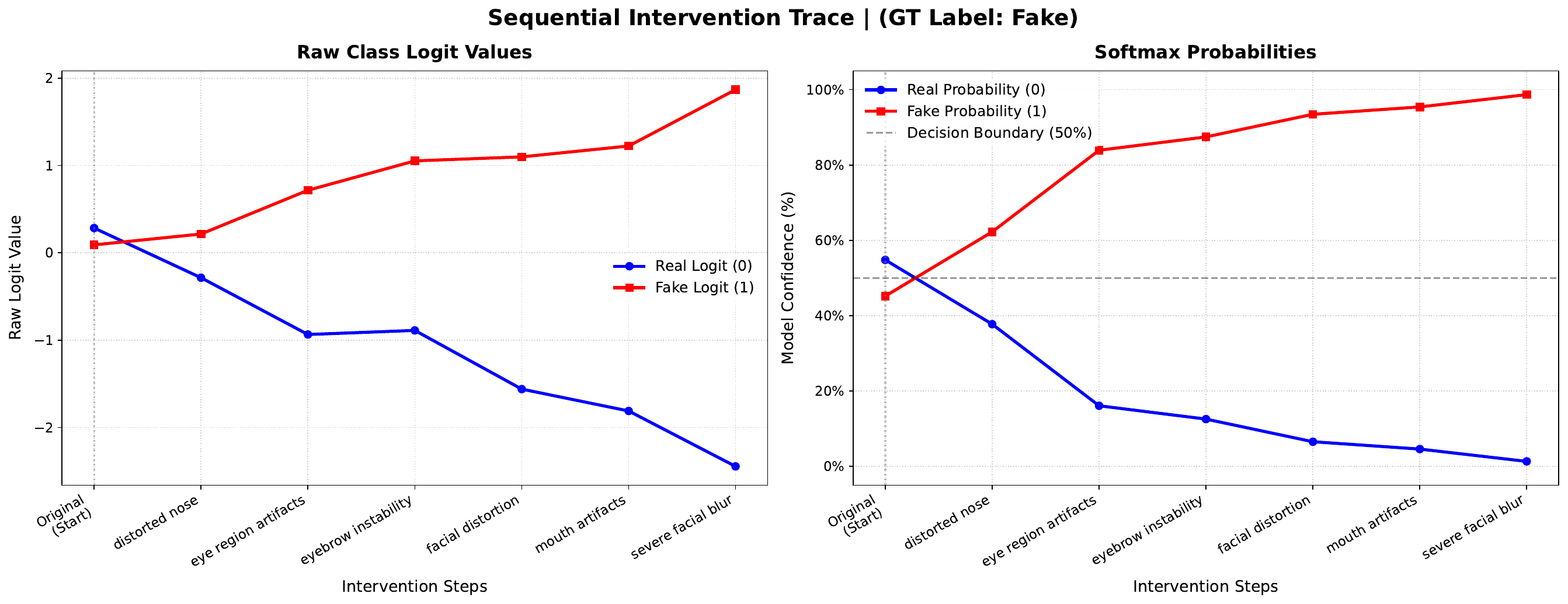}
        \caption{Ground Truth: Fake (Predicted: Real)}
        \label{fig:fake_trace}
    \end{subfigure}
    \caption{Sequential concept intervention traces for two misclassified samples: (a) ground-truth real, (b) ground-truth fake. At each step, a single concept is intervened on while retaining all previous interventions. The left panels show the evolution of the raw class logits, whereas the right panels show the corresponding softmax probabilities.}
    \label{fig:intervention_traces}
\end{figure*}

\paragraph{Intervention traces.}
Fig.~\ref{fig:intervention_traces} illustrates the evolution of the raw class logits and corresponding softmax probabilities during sequential concept interventions for two randomly selected samples. As shown in Fig.~\ref{fig:real_trace}, a ground-truth real sample initially misclassified as fake exhibits little separation between the class logits after the first intervention. Subsequent concept interventions progressively decouple the logits, widening the decision margin and stabilizing the correct prediction. Conversely, Fig.~\ref{fig:fake_trace} shows a ground-truth fake sample initially misclassified as real, where the first intervention immediately shifts the prediction across the decision boundary. Subsequent interventions further increase the target-class logit while suppressing the incorrect-class logit, resulting in steadily increasing prediction confidence.

\section{Conclusion}
\label{sec:conclusion}

In this work, we introduce DF-CBM, a region-aware concept bottleneck model for explainable deepfake detection. To address the gap between visually grounded evidence and language-based explanations, DF-CBM constructs a compact vocabulary of manipulation-related concepts, associates each concept with anatomically meaningful facial and boundary regions and learns a concept-specific masked attention mechanism that grounds concept predictions in localized visual evidence. Through intra-dataset and cross-dataset evaluations, we show that DF-CBM outperforms concept-based baselines in both concept prediction and deepfake classification while remaining competitive with state-of-the-art black-box detectors. Additionally, our qualitative and intervention analyses further provide interpretable evidence linking detected manipulation concepts to the final decision. 

\section*{Acknowledgement}

This research has been supported by the European Commission funded program DETECTOR, under Horizon Europe Grant Agreement 101225942.

{
    \small
    \bibliographystyle{ieeenat_fullname}
    \bibliography{main}

@String(ICASSP=	{ICASSP})

@String(AAAI = {AAAI})

@inproceedings{benou2025show,
  title={Show and tell: Visually explainable deep neural nets via spatially-aware concept bottleneck models},
  author={Benou, Itay and Raviv, Tammy Riklin},
  booktitle={Proceedings of the Computer Vision and Pattern Recognition Conference},
  pages={30063--30072},
  year={2025}
}

@inproceedings{radford2021learning,
  title={Learning transferable visual models from natural language supervision},
  author={Radford, Alec and Kim, Jong Wook and Hallacy, Chris and Ramesh, Aditya and Goh, Gabriel and Agarwal, Sandhini and Sastry, Girish and Askell, Amanda and Mishkin, Pamela and Clark, Jack and others},
  booktitle={International conference on machine learning},
  pages={8748--8763},
  year={2021},
  organization={PmLR}
}

@inproceedings{narayan2025segface,
  title={Segface: Face segmentation of long-tail classes},
  author={Narayan, Kartik and Vs, Vibashan and Patel, Vishal M},
  booktitle={Proceedings of the AAAI Conference on Artificial Intelligence},
  volume={39},
  issue={6},
  pages={6182--6190},
  year={2025}
}

@inproceedings{li2020face,
  title={Face x-ray for more general face forgery detection},
  author={Li, Lingzhi and Bao, Jianmin and Zhang, Ting and Yang, Hao and Chen, Dong and Wen, Fang and Guo, Baining},
  booktitle={Proceedings of the IEEE/CVF conference on computer vision and pattern recognition},
  pages={5001--5010},
  year={2020}
}

@inproceedings{rossler2019faceforensics,
  title={Faceforensics++: Learning to detect manipulated facial images},
  author={Rossler, Andreas and Cozzolino, Davide and Verdoliva, Luisa and Riess, Christian and Thies, Justus and Nie{\ss}ner, Matthias},
  booktitle={Proceedings of the IEEE/CVF international conference on computer vision},
  pages={1--11},
  year={2019}
}

@inproceedings{koh2020concept,
  title={Concept bottleneck models},
  author={Koh, Pang Wei and Nguyen, Thao and Tang, Yew Siang and Mussmann, Stephen and Pierson, Emma and Kim, Been and Liang, Percy},
  booktitle={International conference on machine learning},
  pages={5338--5348},
  year={2020},
  organization={PMLR}
}

@article{espinosa2022concept,
  title={Concept embedding models: Beyond the accuracy-explainability trade-off},
  author={Espinosa Zarlenga, Mateo and Barbiero, Pietro and Ciravegna, Gabriele and Marra, Giuseppe and Giannini, Francesco and Diligenti, Michelangelo and Shams, Zohreh and Precioso, Frederic and Melacci, Stefano and Weller, Adrian and others},
  journal={Advances in neural information processing systems},
  volume={35},
  pages={21400--21413},
  year={2022}
}

@inproceedings{oikarinen2023labelfree,
  title={Label-free Concept Bottleneck Models},
  author={Tuomas Oikarinen and Subhro Das and Lam M. Nguyen and Tsui-Wei Weng},
  booktitle={The Eleventh International Conference on Learning Representations },
  year={2023}
}

@inproceedings{yuksekgonul2023posthoc,
  title={Post-hoc Concept Bottleneck Models},
  author={Mert Yuksekgonul and Maggie Wang and James Zou},
  booktitle={The Eleventh International Conference on Learning Representations},
  year={2023}
}

@article{knab2026whats,
  title={What{\textquoteright}s in the Bottle? A Survey and Roadmap of Concept Bottleneck Models},
  author={Patrick Knab and David Steinmann and Christian Bartelt and Kristian Kersting and Bernt Schiele and Thomas Seidl and Udo Schlegel and Wolfgang Stammer},
  journal={Transactions on Machine Learning Research},
  issn={2835-8856},
  year={2026}
}

@inproceedings{wang2023learning,
  title={Learning bottleneck concepts in image classification},
  author={Wang, Bowen and Li, Liangzhi and Nakashima, Yuta and Nagahara, Hajime},
  booktitle={Proceedings of the ieee/cvf conference on computer vision and pattern recognition},
  pages={10962--10971},
  year={2023}
}

@inproceedings{jeon2024localityaware,
  title={Locality-aware Concept Bottleneck Model},
  author={Sujin Jeon and Inwoo Hwang and Sanghack Lee and Byoung-Tak Zhang},
  booktitle={UniReps: 2nd Edition of the Workshop on Unifying Representations in Neural Models},
  year={2024}
}

@inproceedings{sammani2026clip,
  title={CLIP-Free, Label Free, Unsupervised Concept Bottleneck Models},
  author={Sammani, Fawaz and Fischer, Jonas and Deligiannis, Nikos},
  booktitle={Proceedings of the IEEE/CVF Conference on Computer Vision and Pattern Recognition},
  pages={3262--3272},
  year={2026}
}

@inproceedings{trinh2021interpretable,
  title={Interpretable and trustworthy deepfake detection via dynamic prototypes},
  author={Trinh, Loc and Tsang, Michael and Rambhatla, Sirisha and Liu, Yan},
  booktitle={Proceedings of the IEEE/CVF winter conference on applications of computer vision},
  pages={1973--1983},
  year={2021}
}

@article{bouter2024protoexplorer,
  title={ProtoExplorer: Interpretable forensic analysis of deepfake videos using prototype exploration and refinement},
  author={Bouter, Merel de Leeuw den and Pardo, Javier Lloret and Geradts, Zeno and Worring, Marcel},
  journal={Information Visualization},
  volume={23},
  number={3},
  pages={239--257},
  year={2024},
  publisher={SAGE Publications Sage UK: London, England}
}

@inproceedings{zhang2024common,
  title={Common sense reasoning for deepfake detection},
  author={Zhang, Yue and Colman, Ben and Guo, Xiao and Shahriyari, Ali and Bharaj, Gaurav},
  booktitle={European conference on computer vision},
  pages={399--415},
  year={2024},
  organization={Springer}
}

@inproceedings{hondru2026exddv,
  title={Exddv: A new dataset for explainable deepfake detection in video},
  author={Hondru, Vlad and Hogea, Eduard and Onchis, Darian and Ionescu, Radu Tudor},
  booktitle={Proceedings of the IEEE/CVF Winter Conference on Applications of Computer Vision},
  pages={4273--4284},
  year={2026}
}

@inproceedings{jiang2026tridf,
  title={TriDF: Evaluating Perception, Detection, and Hallucination for Interpretable DeepFake Detection},
  author={Jiang-Lin, Jian-Yu and Huang, Kang-Yang and Zou, Ling and Lo, Ling and Yang, Sheng-Ping and Tseng, Yu-Wen and Lin, Kun-Hsiang and Chen, Chia-Ling and Ta, Yu-Ting and Wang, Yan-Tsung and others},
  booktitle={Proceedings of the IEEE/CVF Conference on Computer Vision and Pattern Recognition},
  pages={17087--17098},
  year={2026}
}

@inproceedings{huang2025sida,
  title={Sida: Social media image deepfake detection, localization and explanation with large multimodal model},
  author={Huang, Zhenglin and Hu, Jinwei and Li, Xiangtai and He, Yiwei and Zhao, Xingyu and Peng, Bei and Wu, Baoyuan and Huang, Xiaowei and Cheng, Guangliang},
  booktitle={Proceedings of the Computer Vision and Pattern Recognition Conference},
  pages={28831--28841},
  year={2025}
}

@inproceedings{soltandoost2025extracting,
  title={Extracting local information from global representations for interpretable deepfake detection},
  author={Soltandoost, Elahe and Plesh, Richard and Schuckers, Stephanie and Peer, Peter and {\v{S}}truc, Vitomir},
  booktitle={Proceedings of the Winter Conference on Applications of Computer Vision},
  pages={1629--1639},
  year={2025}
}

@inproceedings{sun2025towards,
  title={Towards general visual-linguistic face forgery detection},
  author={Sun, Ke and Chen, Shen and Yao, Taiping and Zhou, Ziyin and Ji, Jiayi and Sun, Xiaoshuai and Lin, Chia-Wen and Ji, Rongrong},
  booktitle={Proceedings of the IEEE/CVF Conference on Computer Vision and Pattern Recognition},
  pages={19576--19586},
  year={2025}
}

@inproceedings{vanian2026fake,
  title={Why Fake? Unveiling the Semantic Vocabulary of Deepfake Detectors},
  author={Vanian, Vazgken and Doumanoglou, Alexandros and Zarpalas, Dimitris},
  booktitle={Proceedings of the IEEE/CVF Conference on Computer Vision and Pattern Recognition},
  pages={4101--4110},
  year={2026}
}

@inproceedings{tan2026veritas,
  title={Veritas: Generalizable Deepfake Detection via Pattern-Aware Reasoning},
  author={Hao Tan and jun lan and Zichang Tan and Senyuan Shi and Ajian Liu and Chuanbiao Song and Huijia Zhu and Weiqiang Wang and Jun Wan and Zhen Lei},
  booktitle={The Fourteenth International Conference on Learning Representations},
  year={2026}
}

@inproceedings{xu2025fakeshield,
  title={Fakeshield: Explainable image forgery detection and localization via multi-modal large language models},
  author={Xu, Zhipei and Zhang, Xuanyu and Li, Runyi and Tang, Zecheng and Huang, Qing and Zhang, Jian},
  booktitle={International Conference on Learning Representations},
  volume={2025},
  pages={31186--31216},
  year={2025}
}

@inproceedings{jung2026rich,
  title={A Rich Knowledge Space for Scalable Deepfake Detection},
  author={Jung, Inho and Choi, Hyeongjun and Le, Binh M and Na, Hohyun and Woo, Simon S},
  booktitle={The Fourteenth International Conference on Learning Representations},
  year={2026}
}

@inproceedings{shiohara2022detecting,
  title={Detecting deepfakes with self-blended images},
  author={Shiohara, Kaede and Yamasaki, Toshihiko},
  booktitle={Proceedings of the IEEE/CVF conference on computer vision and pattern recognition},
  pages={18720--18729},
  year={2022}
}

@inproceedings{yan2023ucf,
  title={Ucf: Uncovering common features for generalizable deepfake detection},
  author={Yan, Zhiyuan and Zhang, Yong and Fan, Yanbo and Wu, Baoyuan},
  booktitle={Proceedings of the IEEE/CVF international conference on computer vision},
  pages={22412--22423},
  year={2023}
}

@inproceedings{sun2026dfd,
  title={DFD-HR: Generalizable Deepfake Detection via Hierarchical Routing Learning},
  author={Sun, Jiamu and Yan, Zhiyuan and Zhang, Ke-Yue and Yao, Taiping and Ding, Shouhong},
  booktitle={Proceedings of the IEEE/CVF Conference on Computer Vision and Pattern Recognition},
  pages={13984--13995},
  year={2026}
}

@inproceedings{yan2025orthogonal,
  title={Orthogonal Subspace Decomposition for Generalizable AI-Generated Image Detection},
  author={Yan, Zhiyuan and Wang, Jiangming and Jin, Peng and Zhang, Ke-Yue and Liu, Chengchun and Chen, Shen and Yao, Taiping and Ding, Shouhong and Wu, Baoyuan and Yuan, Li},
  booktitle={International Conference on Machine Learning},
  pages={70268--70288},
  year={2025},
  organization={PMLR}
}

@inproceedings{lin2024fake,
  title={Fake it till you make it: Curricular dynamic forgery augmentations towards general deepfake detection},
  author={Lin, Yuzhen and Song, Wentang and Li, Bin and Li, Yuezun and Ni, Jiangqun and Chen, Han and Li, Qiushi},
  booktitle={European conference on computer vision},
  pages={104--122},
  year={2024},
  organization={Springer}
}

@inproceedings{cui2025forensics,
  title={Forensics adapter: Adapting clip for generalizable face forgery detection},
  author={Cui, Xinjie and Li, Yuezun and Luo, Ao and Zhou, Jiaran and Dong, Junyu},
  booktitle={Proceedings of the Computer Vision and Pattern Recognition Conference},
  pages={19207--19217},
  year={2025}
}

@inproceedings{yang2025d,
  title={D\^{} 3: scaling up deepfake detection by learning from discrepancy},
  author={Yang, Yongqi and Qian, Zhihao and Zhu, Ye and Russakovsky, Olga and Wu, Yu},
  booktitle={Proceedings of the Computer Vision and Pattern Recognition Conference},
  pages={23850--23859},
  year={2025}
}

@inproceedings{koutlis2024leveraging,
  title={Leveraging representations from intermediate encoder-blocks for synthetic image detection},
  author={Koutlis, Christos and Papadopoulos, Symeon},
  booktitle={European Conference on computer vision},
  pages={394--411},
  year={2024},
  organization={Springer}
}

@inproceedings{kashiani2025freqdebias,
  title={FreqDebias: Towards Generalizable Deepfake Detection via Consistency-Driven Frequency Debiasing},
  author={Kashiani, Hossein and Talemi, Niloufar Alipour and Afghah, Fatemeh},
  booktitle={Proceedings of the Computer Vision and Pattern Recognition Conference},
  pages={8775--8785},
  year={2025}
}

@inproceedings{yan2024transcending,
  title={Transcending forgery specificity with latent space augmentation for generalizable deepfake detection},
  author={Yan, Zhiyuan and Luo, Yuhao and Lyu, Siwei and Liu, Qingshan and Wu, Baoyuan},
  booktitle={Proceedings of the IEEE/CVF Conference on Computer Vision and Pattern Recognition},
  pages={8984--8994},
  year={2024}
}

@inproceedings{yang2019exposing,
  title={Exposing deep fakes using inconsistent head poses},
  author={Yang, Xin and Li, Yuezun and Lyu, Siwei},
  booktitle={ICASSP 2019-2019 IEEE international conference on acoustics, speech and signal processing (ICASSP)},
  pages={8261--8265},
  year={2019},
  organization={IEEE}
}

@inproceedings{li2020celeb,
  title={Celeb-df: A large-scale challenging dataset for deepfake forensics},
  author={Li, Yuezun and Yang, Xin and Sun, Pu and Qi, Honggang and Lyu, Siwei},
  booktitle={Proceedings of the IEEE/CVF conference on computer vision and pattern recognition},
  pages={3207--3216},
  year={2020}
}

@article{dolhansky2020deepfake,
  title={The deepfake detection challenge (dfdc) dataset},
  author={Dolhansky, Brian and Bitton, Joanna and Pflaum, Ben and Lu, Jikuo and Howes, Russ and Wang, Menglin and Ferrer, Cristian Canton},
  journal={arXiv preprint arXiv:2006.07397},
  year={2020}
}

@inproceedings{jiang2020deeperforensics,
  title={Deeperforensics-1.0: A large-scale dataset for real-world face forgery detection},
  author={Jiang, Liming and Li, Ren and Wu, Wayne and Qian, Chen and Loy, Chen Change},
  booktitle={Proceedings of the IEEE/CVF conference on computer vision and pattern recognition},
  pages={2889--2898},
  year={2020}
}

@inproceedings{zhou2021face,
  title={Face forensics in the wild},
  author={Zhou, Tianfei and Wang, Wenguan and Liang, Zhiyuan and Shen, Jianbing},
  booktitle={Proceedings of the IEEE/CVF conference on computer vision and pattern recognition},
  pages={5778--5788},
  year={2021}
}

@article{yan2024df40,
  title={Df40: Toward next-generation deepfake detection},
  author={Yan, Zhiyuan and Yao, Taiping and Chen, Shen and Zhao, Yandan and Fu, Xinghe and Zhu, Junwei and Luo, Donghao and Wang, Chengjie and Ding, Shouhong and Wu, Yunsheng and others},
  journal={Advances in Neural Information Processing Systems},
  volume={37},
  pages={29387--29434},
  year={2024}
}

@inproceedings{huang2023implicit,
  title={Implicit identity driven deepfake face swapping detection},
  author={Huang, Baojin and Wang, Zhongyuan and Yang, Jifan and Ai, Jiaxin and Zou, Qin and Wang, Qian and Ye, Dengpan},
  booktitle={Proceedings of the IEEE/CVF conference on computer vision and pattern recognition},
  pages={4490--4499},
  year={2023}
}

@article{cheng2024can,
  title={Can we leave deepfake data behind in training deepfake detector?},
  author={Cheng, Jikang and Yan, Zhiyuan and Zhang, Ying and Luo, Yuhao and Wang, Zhongyuan and Li, Chen},
  journal={Advances in Neural Information Processing Systems},
  volume={37},
  pages={21979--21998},
  year={2024}
}

@inproceedings{ba2024exposing,
  title={Exposing the deception: Uncovering more forgery clues for deepfake detection},
  author={Ba, Zhongjie and Liu, Qingyu and Liu, Zhenguang and Wu, Shuang and Lin, Feng and Lu, Li and Ren, Kui},
  booktitle={Proceedings of the AAAI Conference on Artificial Intelligence},
  volume={38},
  pages={719--728},
  year={2024}
}

@article{luo2023beyond,
  title={Beyond the prior forgery knowledge: Mining critical clues for general face forgery detection},
  author={Luo, Anwei and Kong, Chenqi and Huang, Jiwu and Hu, Yongjian and Kang, Xiangui and Kot, Alex C},
  journal={IEEE Transactions on Information Forensics and Security},
  volume={19},
  pages={1168--1182},
  year={2023},
  publisher={IEEE}
}

@article{tian2024learning,
  title={Learning to discover forgery cues for face forgery detection},
  author={Tian, Jiahe and Chen, Peng and Yu, Cai and Fu, Xiaomeng and Wang, Xi and Dai, Jiao and Han, Jizhong},
  journal={IEEE Transactions on Information Forensics and Security},
  volume={19},
  pages={3814--3828},
  year={2024},
  publisher={IEEE}
}

@article{sun2024diffusionfake,
  title={Diffusionfake: Enhancing generalization in deepfake detection via guided stable diffusion},
  author={Sun, Ke and Chen, Shen and Yao, Taiping and Liu, Hong and Sun, Xiaoshuai and Ding, Shouhong and Ji, Rongrong},
  journal={Advances in Neural Information Processing Systems},
  volume={37},
  pages={101474--101497},
  year={2024}
}

@article{kong2025moe,
  title={Moe-ffd: Mixture of experts for generalized and parameter-efficient face forgery detection},
  author={Kong, Chenqi and Luo, Anwei and Bao, Peijun and Yu, Yi and Li, Haoliang and Zheng, Zengwei and Wang, Shiqi and Kot, Alex C},
  journal={IEEE Transactions on Dependable and Secure Computing},
  year={2025},
  publisher={IEEE}
}

@article{shao2025deepfake,
  title={Deepfake-adapter: Dual-level adapter for deepfake detection},
  author={Shao, Rui and Wu, Tianxing and Nie, Liqiang and Liu, Ziwei},
  journal={International Journal of Computer Vision},
  volume={133},
  number={6},
  pages={3613--3628},
  year={2025},
  publisher={Springer}
}

@inproceedings{fu2025exploring,
  title={Exploring unbiased deepfake detection via token-level shuffling and mixing},
  author={Fu, Xinghe and Yan, Zhiyuan and Yao, Taiping and Chen, Shen and Li, Xi},
  booktitle={Proceedings of the AAAI Conference on Artificial Intelligence},
  volume={39},
  pages={3040--3048},
  year={2025}
}

@inproceedings{esser2024scaling,
  title={Scaling rectified flow transformers for high-resolution image synthesis},
  author={Esser, Patrick and Kulal, Sumith and Blattmann, Andreas and Entezari, Rahim and M{\"u}ller, Jonas and Saini, Harry and Levi, Yam and Lorenz, Dominik and Sauer, Axel and Boesel, Frederic and others},
  booktitle={Forty-first international conference on machine learning},
  year={2024}
}

@article{tian2024visual,
  title={Visual autoregressive modeling: Scalable image generation via next-scale prediction},
  author={Tian, Keyu and Jiang, Yi and Yuan, Zehuan and Peng, Bingyue and Wang, Liwei},
  journal={Advances in neural information processing systems},
  volume={37},
  pages={84839--84865},
  year={2024}
}

@inproceedings{yang2025cogvideox,
  title={Cogvideox: Text-to-video diffusion models with an expert transformer},
  author={Yang, Zhuoyi and Teng, Jiayan and Zheng, Wendi and Ding, Ming and Huang, Shiyu and Xu, Jiazheng and Yang, Yuanming and Hong, Wenyi and Zhang, Xiaohan and Feng, Guanyu and others},
  booktitle={International Conference on Learning Representations},
  volume={2025},
  pages={83048--83077},
  year={2025}
}

@inproceedings{rombach2022high,
  title={High-resolution image synthesis with latent diffusion models},
  author={Rombach, Robin and Blattmann, Andreas and Lorenz, Dominik and Esser, Patrick and Ommer, Bj{\"o}rn},
  booktitle={Proceedings of the IEEE/CVF conference on computer vision and pattern recognition},
  pages={10684--10695},
  year={2022}
}

@inproceedings{karras2018progressive,
  title={Progressive Growing of {GAN}s for Improved Quality, Stability, and Variation},
  author={Tero Karras and Timo Aila and Samuli Laine and Jaakko Lehtinen},
  booktitle={International Conference on Learning Representations},
  year={2018}
}

@article{mirsky2021creation,
  title={The creation and detection of deepfakes: A survey},
  author={Mirsky, Yisroel and Lee, Wenke},
  journal={ACM computing surveys (CSUR)},
  volume={54},
  number={1},
  pages={1--41},
  year={2021},
  publisher={ACM New York, NY, USA}
}

@inproceedings{guo2024animatediff,
  title={AnimateDiff: Animate Your Personalized Text-to-Image Diffusion Models without Specific Tuning},
  author={Yuwei Guo and Ceyuan Yang and Anyi Rao and Zhengyang Liang and Yaohui Wang and Yu Qiao and Maneesh Agrawala and Dahua Lin and Bo Dai},
  booktitle={The Twelfth International Conference on Learning Representations},
  year={2024}
}

@inproceedings{cui2025hallo3,
  title={Hallo3: Highly dynamic and realistic portrait image animation with video diffusion transformer},
  author={Cui, Jiahao and Li, Hui and Zhan, Yun and Shang, Hanlin and Cheng, Kaihui and Ma, Yuqi and Mu, Shan and Zhou, Hang and Wang, Jingdong and Zhu, Siyu},
  booktitle={Proceedings of the Computer Vision and Pattern Recognition Conference},
  pages={21086--21095},
  year={2025}
}

@inproceedings{li2026personalive,
  title={Personalive! expressive portrait image animation for live streaming},
  author={Li, Zhiyuan and Pun, Chi-Man and Fang, Chen and Wang, Jue and Cun, Xiaodong},
  booktitle={Proceedings of the IEEE/CVF Conference on Computer Vision and Pattern Recognition},
  pages={18118--18128},
  year={2026}
}

@inproceedings{zhou2026autocut,
  title={AutoCut: End-to-end advertisement video editing based on multimodal discretization and controllable generation},
  author={Zhou, Milton and Qin, Sizhong and Li, Yongzhi and Chen, Quan and Jiang, Peng},
  booktitle={Proceedings of the IEEE/CVF Conference on Computer Vision and Pattern Recognition},
  pages={37777--37787},
  year={2026}
}

@article{vaccari2020deepfakes,
  title={Deepfakes and disinformation: Exploring the impact of synthetic political video on deception, uncertainty, and trust in news},
  author={Vaccari, Cristian and Chadwick, Andrew},
  journal={Social media+ society},
  volume={6},
  number={1},
  pages={2056305120903408},
  year={2020},
  publisher={SAGE Publications Sage UK: London, England}
}

@article{kirchengast2020deepfakes,
  title={Deepfakes and image manipulation: criminalisation and control},
  author={Kirchengast, Tyrone},
  journal={Information \& Communications Technology Law},
  volume={29},
  number={3},
  pages={308--323},
  year={2020},
  publisher={Taylor \& Francis}
}

@article{wen2026spot,
  title={Spot the fake: Large multimodal model-based synthetic image detection with artifact explanation},
  author={Wen, Siwei and Feng, Peilin and Kang, Hengrui and Wen, Zichen and Chen, Yize and Wu, Jiang and He, Conghui and Li, Weijia and others},
  journal={Advances in Neural Information Processing Systems},
  volume={38},
  pages={58972--59005},
  year={2026}
}

@inproceedings{yan2023deepfakebench,
  title={DeepfakeBench: A Comprehensive Benchmark of Deepfake Detection},
  author={Zhiyuan Yan and Yong Zhang and Xinhang Yuan and Siwei Lyu and Baoyuan Wu},
  booktitle={Thirty-seventh Conference on Neural Information Processing Systems Datasets and Benchmarks Track},
  year={2023}
}

@inproceedings{loshchilov2018decoupled,
  title={Decoupled Weight Decay Regularization},
  author={Ilya Loshchilov and Frank Hutter},
  booktitle={International Conference on Learning Representations},
  year={2019}
}

@misc{dfd2020,
  author       = {{DFD}},
  title        = {{Contributing Data to Deepfake Detection Research}},
  year         = {2020},
  howpublished = {\url{https://ai.googleblog.com/2019/09/contributing-data-to-deepfakedetection.html}},
  note         = {Accessed: 2026-06-30}
}

@inproceedings{li2018ictu,
  title={In ictu oculi: Exposing ai created fake videos by detecting eye blinking},
  author={Li, Yuezun and Chang, Ming-Ching and Lyu, Siwei},
  booktitle={2018 IEEE International workshop on information forensics and security (WIFS)},
  pages={1--7},
  year={2018},
  organization={Ieee}
}

@misc{dfdc2020,
  author       = {{DFDC}},
  title        = {{Deepfake Detection Challenge}},
  year         = {2020},
  howpublished = {\url{https://www.kaggle.com/c/deepfake-detection-challenge}},
  note         = {Accessed: 2026-06-30}
}

@article{vera2025embeddinggemma,
  title={Embeddinggemma: Powerful and lightweight text representations},
  author={Vera, Henrique Schechter and Dua, Sahil and Zhang, Biao and Salz, Daniel and Mullins, Ryan and Panyam, Sindhu Raghuram and Smoot, Sara and Naim, Iftekhar and Zou, Joe and Chen, Feiyang and others},
  journal={arXiv preprint arXiv:2509.20354},
  year={2025}
}

@inproceedings{badr2026frld,
  title={FRLD-DF: Frequency Ring-Guided LoRA Adaptation of DINOv2 Vision Transformer for Generalizable Deepfake Detection},
  author={Badr, Nour Eldin Alaa and Liang, Xing and Nebel, Jean-Christophe and Greenhil, Darrel},
  booktitle={2026 IEEE 20th International Conference on Automatic Face and Gesture Recognition (FG)},
  pages={1--11},
  year={2026},
  organization={IEEE}
}

@article{yang2025qwen3,
  title={Qwen3 technical report},
  author={Yang, An and Li, Anfeng and Yang, Baosong and Zhang, Beichen and Hui, Binyuan and Zheng, Bo and Yu, Bowen and Gao, Chang and Huang, Chengen and Lv, Chenxu and others},
  journal={arXiv preprint arXiv:2505.09388},
  year={2025}
}
}

\end{document}